\documentclass[pdflatex,sn-mathphys-num]{sn-jnl}

\usepackage{graphicx}%
\usepackage{multirow}%
\usepackage{amsmath,amssymb,amsfonts}%
\usepackage{amsthm}%
\usepackage{mathrsfs}%
\usepackage[title]{appendix}%
\usepackage{xcolor}%
\usepackage{textcomp}%
\usepackage{manyfoot}%
\usepackage{booktabs}%
\usepackage{algorithm}%
\usepackage{algorithmicx}%
\usepackage{algpseudocode}%
\usepackage{listings}%
\usepackage{cite}
\usepackage{amsmath,amssymb,amsfonts}
\usepackage{graphicx}
\usepackage{textcomp}
\usepackage{amsthm}
\usepackage{amsmath}
\usepackage{amssymb}
\usepackage{array}
\usepackage{pifont}
\usepackage{wrapfig}
\usepackage{booktabs}
\usepackage{algorithm}
\usepackage{graphicx}
\usepackage{multirow}

\usepackage[table]{xcolor}
\usepackage{pifont}

\definecolor{lightgray}{gray}{0.9}
\definecolor{bestblue}{RGB}{198,219,239}

\newcommand{\best}[1]{\textbf{\textcolor{red}{#1}}}
\newcommand{\second}[1]{\textbf{\textcolor{blue}{#1}}}

\newcommand{\cmark}{\textcolor{green!80!black}{\ding{51}}}
\newcommand{\xmark}{\textcolor{red!85!black}{\ding{55}}}

\theoremstyle{thmstyleone}
\newtheorem{theorem}{Theorem}
\newtheorem{lemma}{Lemma}
\newtheorem{proposition}[theorem]{Proposition}
\newtheorem{corollary}[theorem]{Corollary}

\theoremstyle{thmstyletwo}

\theoremstyle{thmstylethree}
\newtheorem{definition}{Definition}

\begin{document}

\title[A Unified Framework for Skeleton-Based Clinical Assessment]{A Unified Framework with Convex-Concave Decomposition and Sparse Topological Learning for Skeleton-Based Clinical Assessment}


\author[1]{\fnm{Quang-Anh} \sur{N.~D.}}\email{chotala1113@gmail.com}
\author[2]{\fnm{Duc Pham} \sur{Minh}}\email{pmduc2808@gmail.com}
\author[2]{\fnm{Minh-Anh} \sur{Nguyen}}\email{manhmanhh9@gmail.com}
\author[3]{\fnm{Hien Quang} \sur{Kha}}\email{d142111015@tmu.edu.tw}
\author*[4]{\fnm{Manh Hung} \sur{Ha}}\email{hunghm@vnu.edu.vn}
\author[5]{\fnm{Minhhuy} \sur{Le}}\email{huy.leminh@phenikaa-uni.edu.vn}
\author[4,6]{\fnm{Oscal Tzyh-Chiang} \sur{Chen}}\email{oscal@ee.ccu.edu.tw}

\affil*[1]{\orgname{London Digital Twin Research Centre, Middlesex University},
\orgaddress{\city{London}, \postcode{NW4 4BT}, \country{U.K}}}

\affil[2]{\orgname{Agora Intelligence Laboratory},
\orgaddress{\city{Hanoi}, \postcode{100000}, \country{Vietnam}}}

\affil[3]{\orgname{Taipei Medical University},
\orgaddress{\city{Taipei}, \postcode{11031}, \country{Taiwan}}}

\affil*[4]{\orgname{International School, Vietnam National University},
\orgaddress{\city{Hanoi}, \postcode{100000}, \country{Vietnam}}}

\affil[5]{\orgname{Phenikaa University},
\orgaddress{\city{Hanoi}, \postcode{100000}, \country{Vietnam}}}

\affil[6]{\orgname{National Chung Cheng University},
\orgaddress{\city{Chiayi}, \postcode{621301}, \country{Taiwan}}}

\abstract{Skeleton-based rehabilitation assessment faces fundamental limitations of fixed topologies which cannot model compensatory patterns, uniform joint treatments overlook clinically important deviations, and lack of geometric understanding. We propose Concave-Convex Decomposition Sparse Topological Graph Convolutional Network (CC$^2$-STG), a multi-stream model that combines geometric decomposition, sparse feature learning, and adaptive topology. Our framework decomposes joint relationships into complementary convex-concave distance functions to model both local and global movement characteristics, with sparse regularization with proximal optimization selecting clinically relevant geometric features, while multi-scale topological analysis captures non-anatomical joint dependencies in pathological movements. Experiments on the KiMoRe and UIPRMD datasets validate the effectiveness of our approach, achieving state-of-the-art performance on clinical pathological movement assessment.}


\maketitle

\section{Introduction} \label{sec:introduction}
In recent decades, the global demand for physical rehabilitation services has increased dramatically, this increase is largely due to factors such as an aging population and the rising prevalence of non-communicable diseases, particularly musculoskeletal disorders-the leading cause of disability worldwide \citep{1,2}. Despite this growing demand, a substantial gap exists between the supply and demand for rehabilitation services, especially in low and middle-income countries where clinical resources remain limited. The increasing availability of affordable depth sensors has facilitated the development of skeleton-based Action Quality Assessment (AQA) as an automated means of supporting clinical evaluations \citep{zhang-2025,KOURBANE2025109578,10256137} to enhance efficiency, accessibility, and personalized care delivery \citep{4}. Within this context, accurate assessment of movement quality is essential, as clinicians must figure out subtle deviations in posture, joint coordination, and compensatory strategies over time. Motor control theory further suggests that movement quality emerges from the complex interplay of geometric body configurations, differential relationships between joint positions, and the topological structure of the kinematic chain \citep{bernstein1967coordination}. This perspective reinforces the need for computational models capable of capturing nuanced, multi-joint dependencies that reflect real human motor behavior. However, rehabilitation movements are variable, patient-specific, and often characterized by sparse yet clinically meaningful structural abnormalities, posing significant challenges for designing robust and discriminative spatio-temporal representations.

Existing skeleton-based AQA systems have made substantial progress, yet three high-level limitations remain prominent in clinical settings. First, most architectures rely on fixed or manually defined skeletal topologies, which restrict their ability to capture dynamic and non-anatomical dependencies commonly observed in compensatory motions. Second, current representations often lack multi-scale adaptability, limiting their capacity to model both localized geometric irregularities and global postural configurations. Third, many approaches treat all joints or joint pairs uniformly, overlooking the sparsity and structural selectivity inherent to clinically relevant deviations.

\begin{figure}[!ht]
    \centering
    \includegraphics[width=0.6\linewidth]{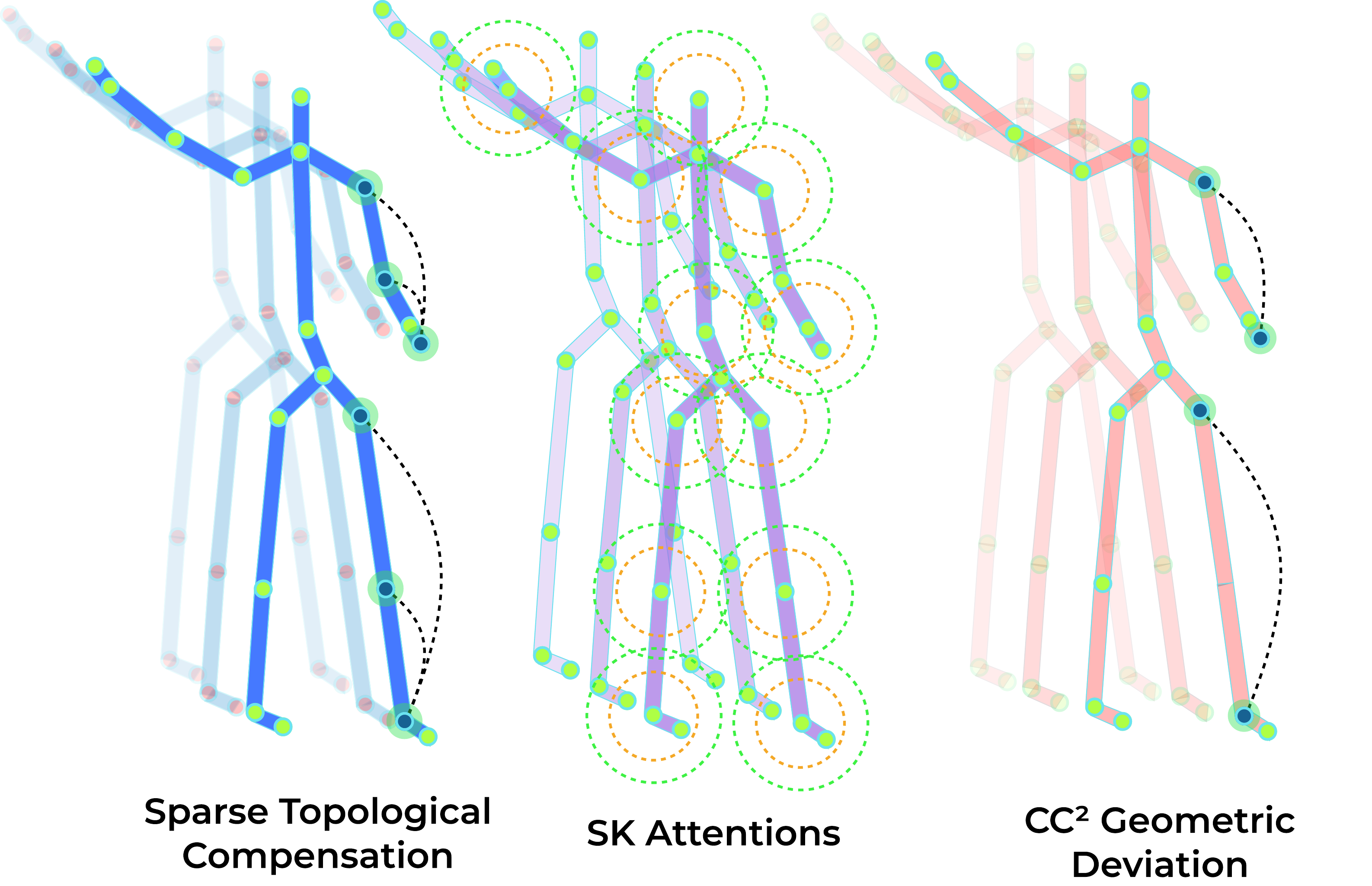}
    \caption{Overview of the proposed CC$^2$-STG pipeline through three parallel streams.}
    \label{fig:abs}
\end{figure}

To address these challenges, we propose Convex-Concave Sparse Topological Graph Convolutional Network (CC$^2$-STG), a unified three-stream framework that integrates multi-scale geometric reasoning, sparse Convex-Concave (CC) decomposition, and adaptive topological modeling. Our approach captures fine-grained local deviations, global posture structure, and non-anatomical relational patterns through complementary spatial streams, further fused with temporal modeling for reliable quality assessment as illustrated in Fig~\ref{fig:abs}. We demonstrate that this design substantially improves robustness and clinical interpretability across diverse rehabilitation scenarios.

\textbf{Our contributions are summarized as follows:}
\begin{itemize}
    \item We introduce CC$^2$-STG a unified AQA framework that jointly models multi-scale geometry, sparse structural decomposition, and adaptive topological relationships.
    \item We propose a Convex-Concave geometric stream with proximal sparsity learning to highlight clinically meaningful joint-pair interactions.
    \item We develop a multi-scale topological module that captures dynamic and non-anatomical joint relations arising in compensatory behaviors.
\end{itemize}

\section{Related Works}
\textbf{Sequence-based skeleton representations.} Early approaches to skeleton-based AQA modeled joint trajectories as sequences or pseudo-images for convolutional and recurrent architectures. Examples include sequence-encoding models such as Liao \citep{rd1} and probabilistic scoring frameworks using Gaussian Mixture Models \citep{WILLIAMS201913,yao2023contrastive}. While effective for basic temporal encoding, these methods rely on normal Euclidean representations and restrictive distributional assumptions that fail to capture the non-Euclidean structure and complex geometric variability of rehabilitation movements.

\noindent\textbf{Graph-based spatio-temporal modeling.} Spatio-temporal graph convolutional networks (ST-GCNs) \citep{10.5555/3298023.3298186,yan2018spatial} explicitly model skeletal anatomy through fixed adjacency structures, and have inspired hierarchical and multi-part variants for AQA \citep{10.1007/s10489-023-04613-5,10.1109/TCSVT.2023.3281413}. However, manually defined adjacency matrices constrain the ability to represent non-anatomical or context-dependent joint relationships, particularly those associated with compensatory patterns in clinical motions. Moreover, many GCN-based systems assume equal importance across joints and edges, limiting their ability to capture sparse, clinically salient deviations.

\noindent\textbf{Transformer-based and hybrid architectures.} Recent transformer-based AQA models, such as PhysioFormer \citep{marusic:hal-04857956} and Interpretable Long-Term AQA \citep{Dong2024InterpretableLA}, exploit long-range temporal attention to capture extended motion patterns. Nonetheless, tokenized skeleton sequences and homogeneous attention struggle to account for multi-scale geometric irregularities and localized joint-level distortions. These limitations reduce the ability of transformers to model the fine-grained structural variability fundamental to rehabilitation assessment.

Across these research directions, current AQA systems face three technical obstacles: reliance on fixed skeletal topology, limited adaptability to multi-scale spatial structure, and insufficient geometric expressiveness for sparse clinical deviations. These persistent limitations motivate the structurally adaptive, sparsity-aware, and geometrically enriched framework proposed in this work.

\section{Methodology}
\subsection{Problem Formulation and Notation}
Using skeletal-based movement quality assessment in clinical rehabilitation presents a fundamental challenge in quantifying the quality of human motion based on temporal skeletal information. We begin by formally defining the skeletal representation and graph structure that underlies our approach. A human skeleton at time $t$ is represented as a graph $\mathcal{G}_t = (\mathcal{V}, \mathcal{E}, \mathbf{F}_t)$, where $\mathcal{V} = \{v_1, v_2, \ldots, v_N\}$ denotes the set of $N$ joints, $\mathcal{E} \subseteq \mathcal{V} \times \mathcal{V}$ represents the natural skeletal connectivity defined by the human kinematic chain, and $\mathbf{F}_t \in \mathbb{R}^{N \times D}$ contains the feature vectors for all joints. Each joint $v_i$ is associated with a feature vector $\mathbf{f}_{t,i} = [\mathbf{q}_{t,i}; \mathbf{p}_{t,i}] \in \mathbb{R}^D$, where $\mathbf{q}_{t,i} \in \mathbb{R}^4$ represents the quaternion orientation and $\mathbf{p}_{t,i} \in \mathbb{R}^3$ denotes the 3D spatial position in the coordinate system. 

\begin{figure*}[!ht]
    \centering
    \includegraphics[width=\linewidth]{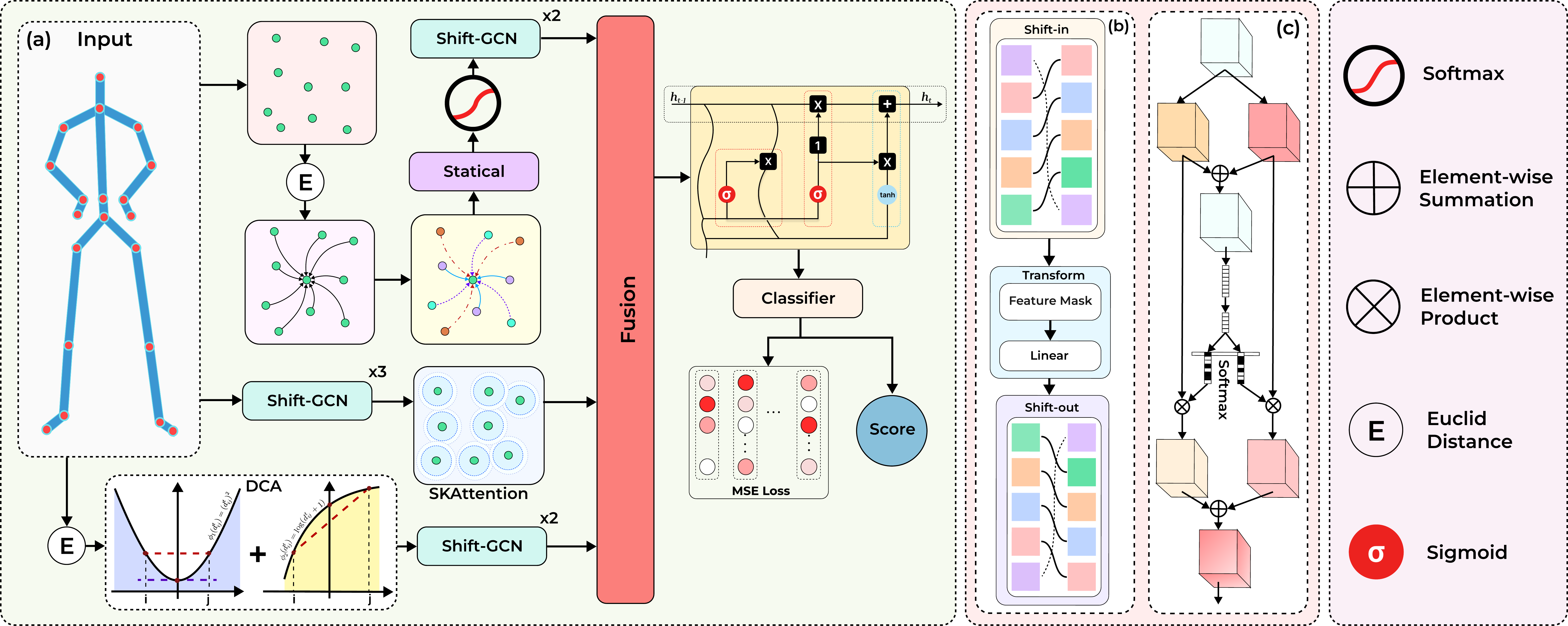}
    \caption{The proposed CC$^2$-STG architecture. (a) Overall architecture with three parallel streams-skeleton-geometry with SK attention, Convex-Concave geometric decomposition, and sparse topological streams-fused and processed through GRU layers for clinical score regression. (b) Shift-GCN block with shift-in and shift-out operations. (c)  SK attention module architecture with multi-scale convolution branches.}
    \label{fig:architecture}
\end{figure*}

The adjacency matrix $\mathbf{A} \in \{0,1\}^{N \times N}$ encodes the skeletal connectivity, where $A_{ij} = 1$ if joints $v_i$ and $v_j$ are connected by a bone, and $A_{ij} = 0$ otherwise. We denote the degree matrix as $\mathbf{D}$, where $D_{ii} = \sum_j A_{ij}$, and we define the normalized adjacency matrix following by~\citep{kipf2017semi} as $\tilde{\mathbf{A}} = \mathbf{D}^{-1/2}\mathbf{A}\mathbf{D}^{-1/2}$. The temporal sequence of skeletons forms a spatial-temporal graph $\mathcal{G} = \{\mathcal{G}_1, \mathcal{G}_2, \ldots, \mathcal{G}_T\}$. Given a temporal sequence of 3D skeleton data $\mathbf{X} = \{\mathbf{x}_1, \mathbf{x}_2, \ldots, \mathbf{x}_T\} \in \mathbb{R}^{T \times N \times D}$, where $T$ denotes the number of temporal frames, $N$ represents the number of skeletal joints, and $D$ is the feature dimension encompassing both joint positions and orientations, our objective is to learn a mapping function $f: \mathbb{R}^{T \times N \times D} \rightarrow \mathbb{R}^K$ that predicts clinical assessment scores $\mathbf{y} \in \mathbb{R}^K$ reflecting movement quality across $K$ evaluation criteria.

\subsection{Overview of the Proposed Approach}
We propose a novel three-stream GNN architecture that addresses the limitations of existing approaches through complementary feature extraction pathways, each designed to capture distinct aspects of movement quality. Our method, termed CC$^2$-STG (Convex-Concave Sparse Topological Graph Convolutional Network), integrates three parallel processing streams: (\textit{i}) a skeleton-geometry stream with Selective Kernel (SK) attention \citep{li2019selective} for adaptive multi-scale spatial feature learning, (\textit{ii}) a Convex-Concave geometric relationship stream with sparse regularization via proximal gradient optimization~\citep{beck2009fast,parikh2014proximal}, and (\textit{iii}) a sparse topological stream that captures local neighborhood structures through multi-scale k-nearest neighbor analysis.

The three streams operate in parallel on the input skeletal sequence, extracting features at different levels of abstraction. These features are subsequently fused through a learnable fusion layer, followed by temporal modeling via Gated Recurrent Units (GRU) to capture the sequential dynamics of movement execution. The entire architecture is trained end-to-end using movement quality scores annotated by clinical experts for the target regression assessment task. The proposed CC$^2$-STG architecture overview is illustrated in Figure~\ref{fig:architecture}.


\subsection{Skeleton-Geometry Processing Stream}
The first stream processes the raw skeletal features through a sequence of spatial-temporal graph convolution layers augmented with adaptive multi-scale feature selection. We employ Shift-GCN~\citep{cheng2020skeleton} as the fundamental building block, which extends standard graph convolution by proposing temporal shift operations to model motion dynamics effectively without additional parameters. To enable adaptive multi-scale spatial feature extraction, we utilized SK attention~\citep{li2019selective} (Illustrated in Fig~\ref{fig:architecture}. The SK attention mechanism dynamically fuses features from multiple parallel convolution branches with different receptive fields, allowing the network to adaptively select the appropriate scale for each spatial location. Specifically, given $M$ parallel convolution branches with kernel sizes $\mathcal{K} = \{k_1, k_2, \ldots, k_M\}$, where we use $\mathcal{K} = \{1, 3, 5\}$, we obtain $M$ feature representations through temporal convolutions along the graph-structured data:

\begin{equation}
\mathbf{U}_m = \text{Conv}_{k_m}\left(\mathbf{X}^{(L)}\right), \quad m = 1, 2, \ldots, M,
\end{equation}
where $\mathbf{X}^{(L)}$ denotes the output from the final Shift-GCN layer. After that, we performed an feature fusion process through a two-stage process by compute global spatial-temporal statistics via global average pooling:

\begin{equation}
\mathbf{s} = \frac{1}{M} \sum_{m=1}^M \frac{1}{\tau N} \sum_{t=1}^\tau \sum_{i=1}^N \mathbf{U}_m(t, i, :),
\end{equation}
where $\mathbf{s} \in \mathbb{R}^{C}$ aggregates information across all branches, time steps, and spatial locations. Subsequently, we compute branch-specific attention weights through a compact feature extraction and expansion process by $\mathbf{z} = \sigma(\mathbf{W}_{\text{fc}}\mathbf{s}), \quad \mathbf{a}_m = \text{softmax}_m(\mathbf{W}_m\mathbf{z})$ where $\mathbf{W}_{\text{fc}} \in \mathbb{R}^{d \times C}$ reduces dimensionality to $d = \max(C/r, L)$ with reduction ratio $r$ and minimum dimension $L$, $\mathbf{W}_m \in \mathbb{R}^{C \times d}$ expands features for branch $m$, and the softmax operation ensures $\sum_{m=1}^M a_{m,c} = 1$ for each channel $c$. The final output combines all branches weighted by their attention:

\begin{equation}
\mathbf{X}_{\text{skel}} = \sum_{m=1}^M \mathbf{a}_m \odot \mathbf{U}_m,
\end{equation}
where $\odot$ denotes element-wise multiplication with broadcasting. This design allows the network to emphasize features from the most informative receptive field scale for each spatial location and temporal instant, which is important for capturing both fine-grained local movements and coarse-grained global postural configurations.

\subsection{Convex-Concave Geometric Relationship Stream}
In this second stream, we propose a geometric feature extraction module motivated by decomposing pairwise joint distances into complementary convex and concave components. We employ computationally efficient proximal gradient optimization~\citep{parikh2014proximal,beck2009fast} for sparse weight learning. This design is motivated by the observation that movement quality assessment involves geometric constraints that benefit from convex decomposition.

The main ideas following by the explicit geometric modeling within the Convex-Concave geometric stream, which decomposes pairwise joint distances into complementary convex and concave components and applies proximal gradient optimization~\citep{parikh2014proximal,beck2009fast} with soft thresholding to learn sparse importance weights. By decomposing distance metrics into squared Euclidean and logarithmic components-functions with complementary growth properties-we enable the model to capture both local fine-grained deviations and global coarse-grained relationships simultaneously. This dual representation is particularly well-suited for clinical movement quality assessment, which fundamentally relies on geometric constraints at multiple scales.

\subsubsection{Convex-Concave Feature Decomposition}

\begin{figure}[!ht]
    \centering
    \includegraphics[width=0.6\linewidth]{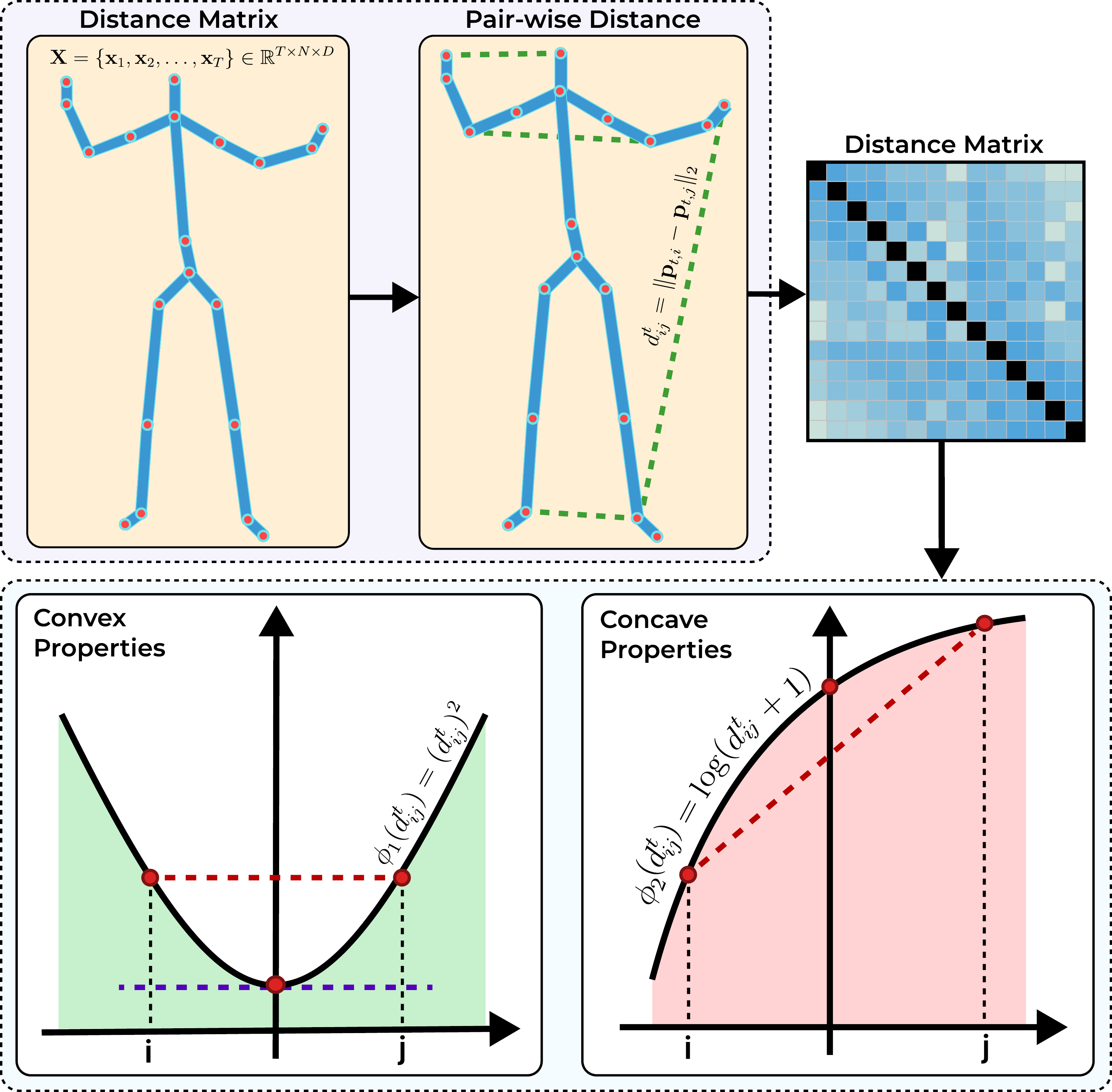}
    \caption{Convex-Concave geometric feature extraction in which skeletal joint positions are used to compute pairwise distances, being decomposed into convex and concave components to capture both local and global movement characteristics.}
    \label{fig:dca}
\end{figure}

Giving 3D joint positions of \textbf{X}, the geometric relationships between joints are fundamentally characterized by pairwise distances. Inspired by DC programming~\citep{tao2005dc}, we decompose geometric features into two complementary functions that capture distinct mathematical properties of spatial relationships, this processing sequence can be showed by Fig~\ref{fig:dca}.

\begin{definition}[Convex-Concave Feature Space]
\label{def:dc_feature_space}
\textit{Let $\mathcal{P} = \{(\mathbf{p}_i, \mathbf{p}_j) : i < j, i,j \in [N]\}$ be the set of all unique joint pairs. We define the convex-concave feature space as the Cartesian product $\Phi = \Phi_1 \times \Phi_2$, where:}
\begin{equation}
\Phi_1 = \{\phi_1(d) : d \in \mathbb{R}_+\}, \quad \Phi_2 = \{\phi_2(d) : d \in \mathbb{R}_+\},
\end{equation}
\textit{with $\phi_1(d) = d^2$ and $\phi_2(d) = \log(d + 1)$ being transformations mapping distances to complementary feature representations.}
\end{definition}

For any pair of joints $(i,j)$, we define the pairwise distance as $d_{ij}^t = \|\mathbf{p}_{t,i} - \mathbf{p}_{t,j}\|_2$. We then construct two feature vectors by applying the transformations:
\begin{equation}
\phi_1(d_{ij}^t) = (d_{ij}^t)^2, \quad \phi_2(d_{ij}^t) = \log(d_{ij}^t + 1),
\end{equation}
where the squared distance emphasizes local geometric deviations with quadratic sensitivity to large distances, while the logarithmic transformation provides robustness to outliers and better captures relative proportions between joint pairs through its compressive nature.


\begin{lemma}[Properties of CC Feature Functions]
\label{lem:dc_properties}
\textit{The following properties of $\phi_1$ and $\phi_2$ justify their use as complementary geometric features:}
\begin{enumerate}
    \item \textit{$\phi_1(d) = d^2$ is strictly convex with $\nabla^2\phi_1(d) = 2 > 0$}
    \item \textit{$\phi_2(d) = \log(d+1)$ is strictly concave with $\nabla^2\phi_2(d) = -\frac{1}{(d+1)^2} < 0$}
    \item \textit{$\phi_1$ exhibits superlinear growth: $\lim_{d \to \infty} \frac{\phi_1(d)}{d} = \infty$}
    \item \textit{$\phi_2$ exhibits sublinear growth: $\lim_{d \to \infty} \frac{\phi_2(d)}{d} = 0$}
    \item \textit{The functions are not affinely related, providing complementary geometric information}
\end{enumerate}
\end{lemma}


\begin{proposition}[Complementary Geometric Sensitivity]
The feature functions $\phi_1(d) = d^2$ and $\phi_2(d) = \log(d+1)$ 
encode complementary distance information. For any $d_a, d_b > 0$ 
with $d_a \neq d_b$:
\begin{equation}
    \frac{\phi_1(d_a)}{\phi_1(d_b)} = \left(\frac{d_a}{d_b}\right)^2 
    \neq 
    \frac{\log(d_a+1)}{\log(d_b+1)} = \frac{\phi_2(d_a)}{\phi_2(d_b)},
\end{equation}
so the two features impose different ordinal information on joint-pair distances. Specifically, $\phi_1$ amplifies large deviations quadratically while $\phi_2$ compresses them logarithmically, providing sensitivity at complementary scales that neither function alone can achieve.
\end{proposition}


By matrix symmetry, we consider only the upper triangular elements of the distance matrix, yielding $M = \frac{N(N-1)}{2}$ unique pairwise features. At time $t$, we construct the CC feature vector, formulated by Eq~\eqref{eq:dcfeat}:
\begin{align}
\boldsymbol{\psi}_t = [&\phi_1(d_{12}^t), \ldots, \phi_1(d_{N-1,N}^t), \nonumber \\
&\phi_2(d_{12}^t), \ldots, \phi_2(d_{N-1,N}^t)]^\top \in \mathbb{R}^{2M},
\label{eq:dcfeat}
\end{align}
representing a concatenation of squared distances and log-distances for all joint pairs. This decomposition naturally separates local and global geometric information while maintaining mathematical tractability for optimization.

\subsubsection{Sparse Feature Weighting via Proximal Optimization}
To learn the importance of different geometric relationships, we utilize learnable weights $\mathbf{w} \in \mathbb{R}^{2M}$ to modulate the CC features, where the weighted feature vector becomes $\tilde{\boldsymbol{\psi}}_t = \mathbf{w} \odot \boldsymbol{\psi}_t$. To prevent overfitting and encourage sparsity in the learned weights, allowing the model to select only the most clinically relevant geometric relationships, we formulate the weight learning as an elastic net regularized optimization problem~\citep{zou2005regularization}:

\begin{equation}
\min_{\mathbf{w}} \mathcal{L}_{\text{task}}(\mathbf{w}) + \lambda \|\mathbf{w}\|_2^2 + \mu \|\mathbf{w}\|_1,
\end{equation}
where $\lambda$ and $\mu$ controls $\ell_2$ and $\ell_1$ regularization strength for smoothness and sparsity, respectively. We employ the computationally efficient proximal gradient 
method~\citep{beck2009fast,parikh2014proximal} with soft thresholding to retains the sparsity-inducing benefits of DC optimization while being compatible with modern DL frameworks and gradient-based training.

\begin{theorem}[Convergence of Proximal Gradient with Soft Thresholding]
\label{thm:proximal_convergence}
Let $\mathcal{L}(\mathbf{w}) = \mathcal{L}_{\text{task}}(\mathbf{w}) + \lambda \|\mathbf{w}\|_2^2 + \mu \|\mathbf{w}\|_1$ be the objective function with $\lambda$ and $\mu$ are larger then zero. Assume $\mathcal{L}_{\text{task}}$ is $L$-smooth (i.e., $\nabla \mathcal{L}_{\text{task}}$ being $L$-Lipschitz continuous), a standard assumption adopted from proximal optimization literature~\citep{beck2009fast,parikh2014proximal} to characterize 
convergence behavior. The proximal gradient update is
\begin{equation}
\label{eq:proximal}
\mathbf{w}^{(k+1)} = \mathcal{S}_{\mu\eta}\left((1-2\lambda\eta)\mathbf{w}^{(k)} 
- \eta \nabla \mathcal{L}_{\text{task}}(\mathbf{w}^{(k)})\right),
\end{equation}
where $\mathcal{S}_{\mu\eta}(x) = \text{sign}(x) \max(|x| - \mu\eta, 0)$ is the soft thresholding operator with threshold $\mu\eta$, converges to a stationary point of $\mathcal{L}(\mathbf{w})$ at the step size satisfying $\eta < \frac{1}{L + 2\lambda}$.
\end{theorem}

\begin{corollary}[Sparsity Guarantee]
\label{cor:sparsity}
\textit{The soft thresholding operator $\mathcal{S}_{\theta}$ induces exact zeros in the weight vector $\mathbf{w}$. Specifically, for any coordinate $i$, if $|(1-2\lambda\eta)w_i^{(k)} - \eta \nabla_i \mathcal{L}_{\text{task}}(\mathbf{w}^{(k)})| < \theta$, then $w_i^{(k+1)} = 0$. The expected number of non-zero weights decreases monotonically with increasing $\mu$.}
\end{corollary}

In practice, we integrate soft thresholding into the forward pass as a stationary-point regularization. At convergence of Eq.~\eqref{eq:proximal}, setting the gradient to zero yields a learning-rate-independent threshold $\frac{\mu}{2\lambda}$:
\begin{equation}
w_i^{\text{soft}} = \text{sign}(w_i) \cdot \max\left(|w_i| - 
\frac{\mu}{2\lambda + \epsilon}, 0\right),
\end{equation}
where $\epsilon = 10^{-8}$ is a small constant for numerical stability. This operator is applied in the forward pass independently of the optimization update in Eq.~\eqref{eq:proximal}, enforcing sparsity while remaining compatible with end-to-end backpropagation, thereby selecting clinically relevant geometric features (see Appendix H for visualization of the learned 
importance scores).

\subsubsection{Feature Normalization and Graph Convolution}
After CC feature extraction and sparse weighting, we apply layer-wise feature normalization to stabilize training by $\hat{\boldsymbol{\psi}}_t = \tilde{\boldsymbol{\psi}}_t - \boldsymbol{\mu}_{\psi} / \boldsymbol{\sigma}_{\psi} + \epsilon$ where $\boldsymbol{\mu}_{\psi}$ and $\boldsymbol{\sigma}_{\psi}$ are the mean and standard deviation computed across both batch and temporal dimensions. These features are then projected to a higher-dimensional space through a MLP with layer norm by:

\begin{align}
\mathbf{h}_t^{\text{CC}} = \text{MLP}_{\text{CC}}(\hat{\boldsymbol{\psi}}_t) = \mathbf{W}_2 \cdot \sigma(\text{LN}(\mathbf{W}_1\hat{\boldsymbol{\psi}}_t + \mathbf{b}_1)) + \mathbf{b}_2,
\end{align}
where $\mathbf{W}_1 \in \mathbb{R}^{2H \times 2M}$ and $\mathbf{W}_2 \in \mathbb{R}^{H \times 2H}$ are learnable projection matrices, $\text{LN}(\cdot)$ denotes layer normalization, and $H$ is the hidden dimension. This produces a temporal sequence of CC features $\mathbf{H}^{\text{CC}} = \{\mathbf{h}_1^{\text{CC}}, \ldots, \mathbf{h}_\tau^{\text{CC}}\} \in \mathbb{R}^{\tau \times H}$. To enable graph-based spatial processing of these global geometric features, we expand the temporal features across the joint dimension and apply Shift-GCN layers by Eq~\eqref{eq:dcgnn}.

\begin{equation}
\mathbf{X}_{\text{CC}} = \text{Shift-GCN}_2(\text{Shift-GCN}_1(\text{Expand}(\mathbf{H}^{\text{CC}}))),
\label{eq:dcgnn}
\end{equation}
where $\text{Expand}: \mathbb{R}^{\tau \times H} \to \mathbb{R}^{\tau \times N \times H}$ replicates the features across all $N$ joints, creating tensors suitable for graph convolution that propagates the global geometric information through the skeletal structure.

\subsection{Sparse Topological Neighborhood Stream}
\begin{figure}[!ht]
    \centering
    \includegraphics[width=0.6\linewidth]{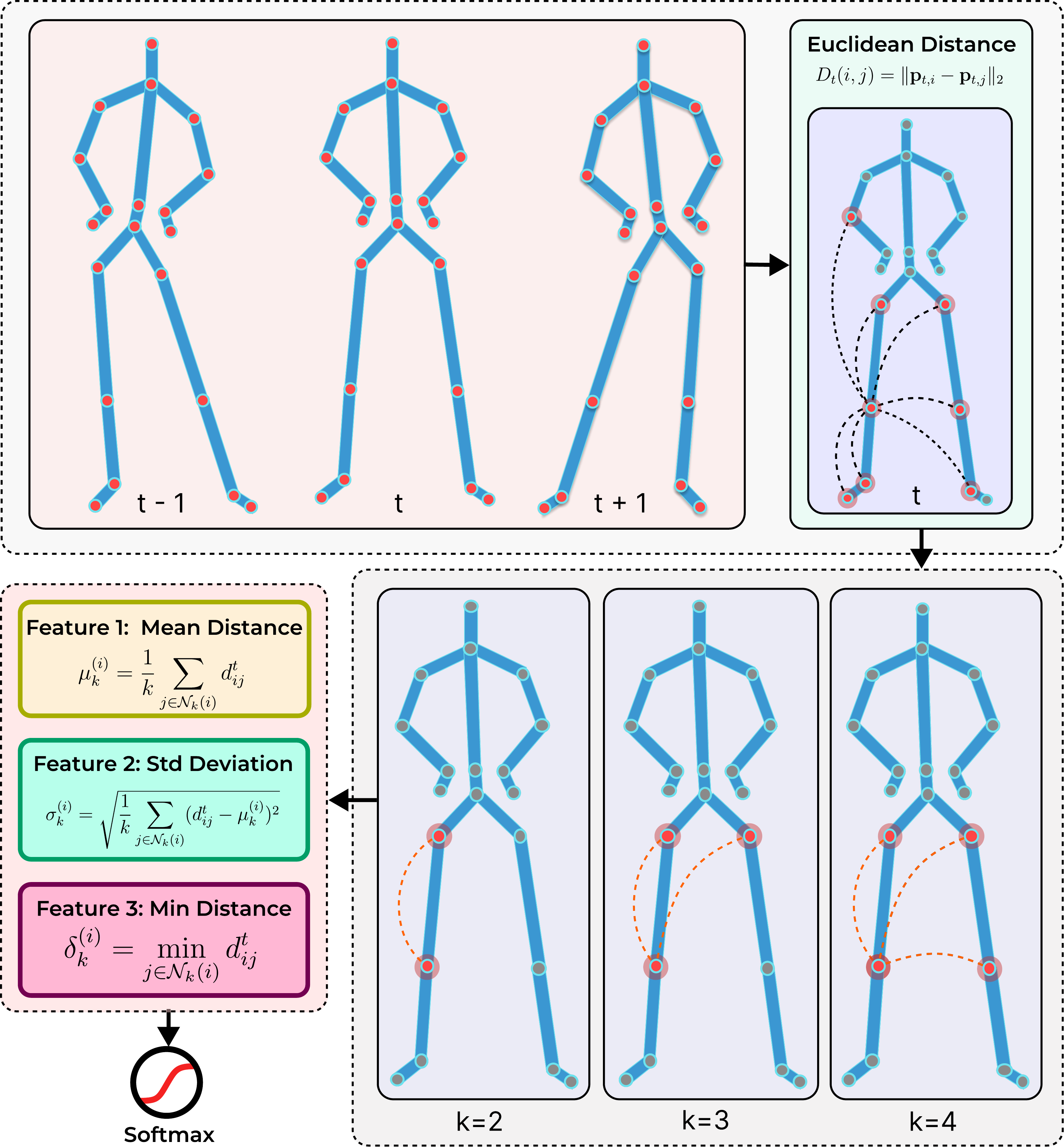}
    \caption{Spare topological feature extraction pipeline. Pairwise Euclidean distances are computed between joints across temporal frames. For each joint, local descriptors are extracted at multiple scales and combined via learned softmax-normalized weights for non-anatomical joint dependencies in compensatory movements.}
    \label{fig:placeholder2}
\end{figure}

The third stream captures local topological structures by connecting hyper structure relationships in the positional space. Unlike the skeleton connectivity $\mathcal{E}$ defined by anatomical constraints, k-Nearest Neighbor (k-NN) topology reflects the actual spatial configuration and can reveal abnormal postures or compensatory movements where joints that are typically distant become spatially proximate~\citep{bronstein2017geometric}, this stream operation is illustrated in Fig~\ref{fig:placeholder2}.

\subsubsection{Multi-scale k-NN Feature Extraction}
For each temporal frame $t$, we compute the pairwise Euclidean distance matrix $\mathbf{D}_t \in \mathbb{R}^{N \times N}$ between all joint positions, where $D_t(i,j) = \|\mathbf{p}_{t,i} - \mathbf{p}_{t,j}\|_2$. For each joint $i$, we identify its $k$ nearest neighbors for their values of $k \in \mathcal{K} = \{k_1, k_2, k_3\}$, where typically $k_1 < k_2 < k_3$ (\textit{e.g.}, $\mathcal{K} = \{2, 3, 4\}$) to capture multi-scale neighborhood structures ranging from close neighbors to broader local ones.

For each scale $k \in \mathcal{K}$ and joint $i$, let $\mathcal{N}_k(i) = \{j_1, j_2, \ldots, j_k\}$ denote the set of $k$ nearest neighbors. We compute three statistical features characterizing the local topology:
\begin{align}
\mu_k^{(i)} &= \frac{1}{k}\sum_{j \in \mathcal{N}_k(i)} d_{ij}^t, \label{eq:knn_mean}\\
\sigma_k^{(i)} &= \sqrt{\frac{1}{k}\sum_{j \in \mathcal{N}_k(i)} (d_{ij}^t - \mu_k^{(i)})^2}, \label{eq:knn_std}\\
\delta_k^{(i)} &= \min_{j \in \mathcal{N}_k(i)} d_{ij}^t. \label{eq:knn_min}
\end{align}

The mean distance $\mu_k^{(i)}$ reflects the overall spatial extent of the neighborhood, the standard deviation $\sigma_k^{(i)}$ captures the uniformity of neighbor distribution (low values indicating clustered neighbors, high values suggesting dispersed configurations), and the minimum distance $\delta_k^{(i)}$ identifies the closest spatial relationship.

\begin{definition}[Local Topological Descriptor]
\label{def:topological_descriptor}
\textit{For a joint $i$ at time $t$, we define the multi-scale topological descriptor as the concatenation:}
\begin{align}
\mathbf{f}_t^{(i,\text{topo})} = [&\mu_{k_1}^{(i)}, \sigma_{k_1}^{(i)}, \delta_{k_1}^{(i)}, \mu_{k_2}^{(i)}, \sigma_{k_2}^{(i)}, \delta_{k_2}^{(i)}, \nonumber \\
&\mu_{k_3}^{(i)}, \sigma_{k_3}^{(i)}, \delta_{k_3}^{(i)}]^\top \in \mathbb{R}^{3|\mathcal{K}|},
\end{align}
\textit{which characterizes the local topological context at multiple spatial scales.}
\end{definition}

\begin{lemma}[Stability of k-NN Features]
\label{lem:knn_stability}
\textit{Let $\mathbf{P}_t$ and $\mathbf{P}_t'$ be two configurations of joint positions such that $\|\mathbf{p}_{t,i} - \mathbf{p}_{t,i}'\|_2 < \delta$ for all $i \in [N]$. Under such condition, the k-NN features are Lipschitz continuous with respect to position perturbations:}
\begin{equation}
|\mu_k^{(i)}(\mathbf{P}_t) - \mu_k^{(i)}(\mathbf{P}_t')| \leq 2\delta, \quad |\delta_k^{(i)}(\mathbf{P}_t) - \delta_k^{(i)}(\mathbf{P}_t')| \leq 2\delta.
\end{equation}
\end{lemma}

\begin{theorem}[Local Topological Preservation]
\label{thm:topological_preservation}
\textit{The multi-scale k-NN descriptor $\mathbf{f}_t^{(i,\text{topo})}$ preserves local topological structure. If two configurations $\mathbf{P}_t$ and $\mathbf{P}_t'$ satisfy $\mathbf{f}_t^{(i,\text{topo})}(\mathbf{P}_t) = \mathbf{f}_t^{(i,\text{topo})}(\mathbf{P}_t')$ for all $i \in [N]$ and all $k \in \mathcal{K}$, then the local neighborhoods $\mathcal{N}_k(i)$ have identical distance distributions. This ensures that local geometric relationships relevant for movement quality assessment are preserved.}
\end{theorem}

\subsubsection{Learnable Scale Importance Weighting}
For the model to adapt to the relative importance of different neighborhood scales-recognizing that different movement types require different levels of spatial context-we utilize learnable importance weights $\boldsymbol{\alpha} = [\alpha_1, \alpha_2, \alpha_3]^\top$ implemented as trainable parameters. Additionally, a softmax constraint is applied to ensure valid probability weights by $\alpha_k = \frac{\exp(\tilde{\alpha}_k)}{\sum_{k' \in \mathcal{K}} \exp(\tilde{\alpha}_{k'})}$ and $\sum_{k \in \mathcal{K}} \alpha_k = 1$ where $\tilde{\boldsymbol{\alpha}} \in \mathbb{R}^{|\mathcal{K}|}$ are the unconstrained parameters. The weighted topological features become:

\begin{align}
\mathbf{f}_t^{(i,\text{topo})} = \sum_{j=1}^{|\mathcal{K}|} \alpha_j \cdot [\mu_{k_j}^{(i)}, \sigma_{k_j}^{(i)}, \delta_{k_j}^{(i)}]^\top.
\end{align}

These per-joint features are then processed through Shift-GCN layers to capture spatial dependencies in the topological space by $\mathbf{X}_{\text{topo}} = \text{Shift-GCN}_2(\text{Shift-GCN}_1(\mathbf{F}_{\text{topo}}))$, where $\mathbf{F}_{\text{topo}} \in \mathbb{R}^{\tau \times N \times 9}$ aggregates the topological features across time. The output is spatially pooled via global average pooling over the joint dimension for temporal representations:
\begin{equation}
\mathbf{h}_t^{\text{topo}} = \frac{1}{N}\sum_{i=1}^N \mathbf{X}_{\text{topo}}(t, i, :) \in \mathbb{R}^{H}.
\end{equation}

\subsection{Multi-Stream Fusion and Temporal Modeling}
The three streams produce complementary feature representations of skeleton-geometry features $\mathbf{X}_{\text{skel}} \in \mathbb{R}^{\tau \times N \times C}$, Convex-Concave geometric features $\mathbf{X}_{\text{CC}} \in \mathbb{R}^{\tau \times N \times H}$, and topological features $\mathbf{H}^{\text{topo}} \in \mathbb{R}^{\tau \times H}$. After that, we reshape the skeleton and CC features and concatenate all three streams:

\begin{equation}
\mathbf{X}_{\text{cat}} = [\omega(\mathbf{X}_{\text{skel}}); \omega(\mathbf{X}_{\text{CC}}); \mathbf{H}^{\text{topo}}] \in \mathbb{R}^{\tau \times D_{\text{cat}}},
\end{equation}
where $\omega$ denotes flattening operations and $D_{\text{cat}} = N \cdot C + N \cdot H + H$ is the concatenated dimension. These multi-stream features are fused through a learnable projection network:


\begin{align}
\mathbf{H}_{\text{fused}} = \mathbf{W}_{\text{fuse2}} \cdot \sigma(\varphi(\text{LN}(\mathbf{W}_{\text{fuse1}}\mathbf{X}_{\text{cat}} + \mathbf{b}_1))) + \mathbf{b}_2,
\end{align}
where $\varphi$ represents the Dropout layer, and $\mathbf{W}_{\text{fuse1}} \in \mathbb{R}^{2H \times D_{\text{cat}}}$ and $\mathbf{W}_{\text{fuse2}} \in \mathbb{R}^{H \times 2H}$ are learnable fusion weights that reduce dimensionality to $H$ while enabling non-linear interaction between streams. The fused features $\mathbf{H}_{\text{fused}} \in \mathbb{R}^{\tau \times H}$ capture comprehensive spatial characteristics at each time step. To model the temporal dynamics of movement, we employ GRU layers with $\mathbf{h}_t = \text{GRU}(\mathbf{H}_{\text{fused}}(t,:), \mathbf{h}_{t-1})$, in which $\mathbf{h}_t \in \mathbb{R}^H$ represents the hidden state at time $t$. The GRU processes the sequence from $t=1$ to $t=\tau$, accumulating temporal context through its gating mechanisms. The final hidden state $\mathbf{h}_\tau$ encapsulates the entire sequence's temporal dynamics and is passed through a linear output layer:
\begin{equation}
\hat{\mathbf{y}} = \mathbf{W}_{\text{out}}\mathbf{h}_\tau + \mathbf{b}_{\text{out}},
\end{equation}
where $\mathbf{W}_{\text{out}} \in \mathbb{R}^{K \times H}$ and $\mathbf{b}_{\text{out}} \in \mathbb{R}^K$ produce the predicted assessment scores $\hat{\mathbf{y}} \in \mathbb{R}^K$ for $K$ clinical evaluation criteria.

\section{Experimental Environment}
\noindent\textbf{Datasets.} In this research, we utilized benchmarks of physical rehabilitation datasets, KiMoRe and UIPRMD \citep{kimore,Vakanski2018}, designed to support the assessment of human movement quality. The KiMoRe dataset contains motion recordings from 78 subjects, including 34 patients diagnosed with conditions such as Parkinson’s disease, stroke, and chronic back pain. In contrast, the UIPRMD dataset contains nearly 1,000 motion samples collected from 10 subjects performing 10 rehabilitation exercises under expert supervision to ensure standardized execution.

\noindent\textbf{Loss Functions.} The model is trained end-to-end using a simple Mean Squared Error (MSE) loss between predicted and ground-truth clinical scores:

\begin{equation}
\mathcal{L}_{\text{MSE}} = \frac{1}{B}\sum_{b=1}^B |\hat{\mathbf{y}}_b - \mathbf{y}_b|_2^2,
\end{equation}
where $B$ is the batch size. The total loss includes an elastic net regularization applied exclusively to the Convex-Concave stream weights $\mathbf{w}$ to encourage sparse selection of clinically relevant geometric features. Therefore the total loss function:


\begin{equation}
\mathcal{L}_{\text{total}} = \mathcal{L}_{\text{MSE}} + \lambda \|\mathbf{w}\|_2^2 + \mu \|\mathbf{w}\|_1
\label{eq:tls}
\end{equation}

\noindent\textbf{Implementation.} All experiments were conducted on an singe NVIDIA GeForce RTX 3060 GPU. Training used the AdamW optimizer, elastic-net regularization was applied to the Convex-Concave stream weights with coefficients $\lambda = 10^{-3}$ ($\ell_2$) and $\mu = 10^{-2}$ ($\ell_1$), and a dropout rate of $0.15$ was used throughout the network. The SK attention kernels were configured following the original setting~\citep{li2019selective}, and the $k$-NN topological stream operated at scales $\{2, 3, 4\}$. Each model was trained with a batch size of $32$ for up to $1500$ epochs.





\section{Experiments Results}
\subsection{Evaluation Results}
\noindent\textbf{UIPRMD dataset results.} Table \ref{tab:performance_ui_prmd} reports the MAD comparison on the UIPRMD dataset. In contrast to KiMoRe, the absolute errors on UIPRMD are substantially smaller, ranging only from a few thousandths to a few hundredths, indicating that this dataset is less challenging and is likely reaching saturation. As a result, even earlier methods such as the work of Zhang and collaborators from 2020 continue to secure several top scores, while more recent approaches offer only limited improvements. CC$^2$-STG attains the best results in Exercises 3, 7, and 8, and achieves second-best performance in Exercises 1, 2, 4, and 5. However, the model performs noticeably worse in Exercise 10, where the error is higher than in most baseline methods. These observations suggest that while CC$^2$-STG generalizes effectively across standardized movements, certain exercise categories in UIPRMD-especially those with minimal geometric variation or low temporal complexity may require additional task-specific modeling strategies.

\noindent\textbf{KiMoRe dataset results.} For the KiMoRe dataset, Table~\ref{tab:performance_ex1to5} shows that CC$^2$-STG delivers the strongest overall performance across most exercises and evaluation metrics. With respect to MAD, the model ranks first in Exercises 1, 3, 4, and 5, and second in Exercise 2, demonstrating robust stability across diverse movements. Beyond MAD, the model also provides notable improvements in RMSE and MAPE. It achieves the lowest RMSE in Exercises 1 and 5, and remains the second-best performer in Exercises 2 and 3. For MAPE, CC$^2$-STG consistently outperforms all competing approaches across all five exercises. These coherent advantages across three metrics indicate that the proposed architecture not only reduces absolute deviations but also enhances relative error behavior and temporal magnitude consistency. Although Yao and colleagues obtain the lowest MAD in Exercise 2, their corresponding RMSE and MAPE values remain inferior on the remaining tasks. Averaged over all exercises, CC$^2$-STG achieves the best overall results, confirming that its multi-scale geometric modeling provides robustness results.

\begin{table*}[!t]
\centering
\caption{Performance comparison (MAD $\downarrow$) on UIPRMD Exercises 1--10. Best results in \textbf{\textcolor{red}{red}} and second-best in \textbf{\textcolor{blue}{blue}}.}
\label{tab:performance_ui_prmd}
\resizebox{\textwidth}{!}{
\begin{tabular}{lcccccccccc}
\toprule
\textbf{Model} & Ex1 & Ex2 & Ex3 & Ex4 & Ex5 & Ex6 & Ex7 & Ex8 & Ex9 & Ex10 \\
\midrule
Li \textit{et al.} \citep{li2018cooccurrence} 
& 0.011 & 0.029 & 0.036 & 0.014 & 0.017 & 0.019 & 0.027 & 0.025 & 0.027 & 0.047 \\

Song \textit{et al.} \citep{song2020ra-gcn} 
& 0.011 & 0.006 & \second{0.010} & 0.014 & 0.013 & 0.009 & 0.017 & 0.017 & 0.008 & 0.038 \\

Zhang \textit{et al.} \citep{zhang2020sgn} 
& 0.022 & \best{0.003} & 0.016 & 0.016 & \best{0.003} & \best{0.003} & 0.021 & 0.025 & 0.027 & 0.066 \\

Liao \textit{et al.} \citep{liao2020rehab} 
& 0.011 & 0.028 & 0.029 & 0.012 & 0.019 & 0.012 & 0.018 & 0.023 & 0.023 & 0.042 \\

S. Deb \textit{et al.} \citep{deb2022gcn}
& \second{0.009} & 0.006 & 0.013 & \second{0.006} & 0.008 & 0.006 & 0.011 & 0.016 & 0.008 & 0.031 \\

Mourchid \textit{et al.} \citep{mourchid2023d} 
& 0.011 & 0.009 & 0.013 & 0.009 & 0.009 & 0.013 & 0.022 & 0.020 & 0.013 & 0.014 \\

Yao \textit{et al.} \citep{10256137} 
& 0.013 & 0.012 & 0.014 & 0.009 & 0.007 & 0.010 & 0.018 & \second{0.011} & 0.040 & 0.014 \\

Yao \textit{et al.} \citep{yao2023contrastive} 
& 0.015 & 0.012 & 0.015 & 0.008 & 0.009 & 0.010 & 0.011 & 0.018 & 0.010 & 0.044 \\

Sardari \textit{et al.} \citep{sardari2024lightpra} 
& 0.014 & 0.007 & 0.011 & \second{0.006} & 0.008 & 0.006 & \second{0.010} & \second{0.011} & 0.008 & 0.038 \\

PhysioFormer \citep{marusic:hal-04857956} 
& 0.010 & 0.009 & 0.011 & 0.009 & 0.008 & 0.012 & 0.019 & 0.018 & 0.013 & \second{0.013} \\

FTF-HGCN \citep{zhang-2025} 
& 0.008 & 0.005 & 0.011 & \best{0.004} & \second{0.006} & 0.006 & \best{0.009} & \best{0.010} & \second{0.007} & \best{0.010} \\

STGCN-Seq \citep{KOURBANE2025109578} 
& \best{0.006} & 0.008 & \best{0.009} & \second{0.006} & \best{0.003} & \second{0.004} & 0.009 & 0.013 & \best{0.006} & 0.028 \\

\rowcolor{bestblue}
\textbf{CC$^2$-STG (Ours)}
& \second{0.009} & \second{0.004} & \best{0.009} & \second{0.006} & \second{0.006} & 0.006 & \best{0.009} & \best{0.010} & 0.008 & 0.041 \\
\bottomrule
\end{tabular}
}
\end{table*}

\begin{table*}[t]
\centering
\caption{Performance comparison on the KiMoRe dataset across Exercises 1-5. 
All metrics report lower values as better performance. Best results per column are highlighted in \textbf{\textcolor{red}{red}} and second-best in \textbf{\textcolor{blue}{blue}}.}
\label{tab:performance_ex1to5}

\resizebox{\textwidth}{!}{
\begin{tabular}{l *{5}{c} *{5}{c} *{5}{c}}
\toprule
\multirow{2}{*}{\textbf{Model}} 
& \multicolumn{5}{c}{\textbf{MAD} $\downarrow$}
& \multicolumn{5}{c}{\textbf{RMSE} $\downarrow$}
& \multicolumn{5}{c}{\textbf{MAPE} $\downarrow$} \\
\cmidrule(lr){2-6} \cmidrule(lr){7-11} \cmidrule(lr){12-16}
& Ex1 & Ex2 & Ex3 & Ex4 & Ex5
& Ex1 & Ex2 & Ex3 & Ex4 & Ex5
& Ex1 & Ex2 & Ex3 & Ex4 & Ex5 \\
\midrule

Li \textit{et al.} \citep{li2018cooccurrence} 
& 1.378 & 1.877 & 1.452 & 0.675 & 1.662
& 2.344 & 2.823 & 2.034 & 1.078 & 2.575
& 3.491 & 5.298 & 4.188 & 2.130 & 3.752 \\

Yan \textit{et al.} \citep{yan2018stgcn} 
& 0.889 & 2.096 & 0.604 & 0.842 & 1.218
& 2.017 & 3.262 & 0.799 & 1.331 & 1.951
& 2.339 & 6.136 & 1.727 & 2.325 & 3.802 \\

Liao \textit{et al.} \citep{liao2020rehab} 
& 1.141 & 1.528 & 0.845 & 0.468 & 0.847
& 2.534 & 3.738 & 1.640 & 0.792 & 1.941
& 2.589 & 3.976 & 2.023 & 1.354 & 2.312 \\

Song \textit{et al.} \citep{song2020ra-gcn} 
& 0.977 & 1.282 & 1.033 & 0.715 & 1.536
& 2.165 & 3.345 & 1.679 & 2.018 & 3.198
& 2.605 & 5.202 & 2.968 & 3.599 & 4.959 \\

Zhang \textit{et al.} \citep{zhang2020sgn} 
& 1.757 & 3.139 & 1.737 & 1.202 & 1.853
& 2.916 & 4.140 & 2.615 & 1.836 & 2.916
& 5.034 & 10.456 & 5.774 & 3.797 & 6.531 \\

S. Deb \textit{et al.} \citep{deb2022gcn} 
& 0.799 & 0.774 & 0.369 & 0.347 & 0.621
& 2.024 & 2.130 & 0.856 & 0.644 & 1.181
& 1.926 & 1.472 & 0.728 & 1.222 & 1.591 \\

Yao \textit{et al.} \citep{10256137} 
& 0.444 & \best{0.030} & 0.142 & 0.121 & 0.292
& 0.569 & 0.390 & \best{0.180} & \best{0.148} & 0.378
& 1.105 & 0.864 & 0.437 & \second{0.341} & 0.808 \\

Jleli \textit{et al.} \citep{Jleli2024ArtificialIR}
& 0.482 & 0.521 & 0.389 & 0.478 & 0.489
& 1.026 & 1.102 & 1.206 & 1.054 & 0.996
& 1.112 & 1.205 & 1.098 & 0.981 & 1.108 \\

RAST-G \citep{lim-2025}
& \second{0.225} & 0.227 & 0.231 & 0.221 & \second{0.220}
& \second{0.267} & \best{0.268} & 0.274 & 0.264 & \second{0.264}
& \second{0.364} & \second{0.370} & \second{0.385} & 0.346 & \second{0.349} \\

FTF-HGCN \citep{zhang-2025}
& 0.622 & 0.491 & 0.206 & \second{0.204} & 0.390
& 1.378 & 0.748 & 0.398 & 0.515 & 0.698
& 1.508 & 0.952 & 0.536 & 0.483 & 1.113 \\

STGCN-Seq \citep{KOURBANE2025109578} 
& 0.543 & 0.511 & 0.213 & \second{0.204} & 0.488
& 1.492 & 1.124 & 0.337 & \second{0.218} & 0.724
& 1.362 & 0.766 & 0.620 & 0.514 & 1.412 \\

\rowcolor{bestblue}
\textbf{CC$^2$-STG (Ours)} 
& \best{0.066} & \second{0.105} & \best{0.100} & \best{0.085} & \best{0.099}
& \best{0.118} & \second{0.356} & \second{0.269} & 0.307 & \best{0.207}
& \best{0.160} & \best{0.322} & \best{0.260} & \best{0.258} & \best{0.229} \\
\bottomrule
\end{tabular}}
\end{table*}

\begin{figure*}[!ht]
    \centering
    \includegraphics[width=1\linewidth]{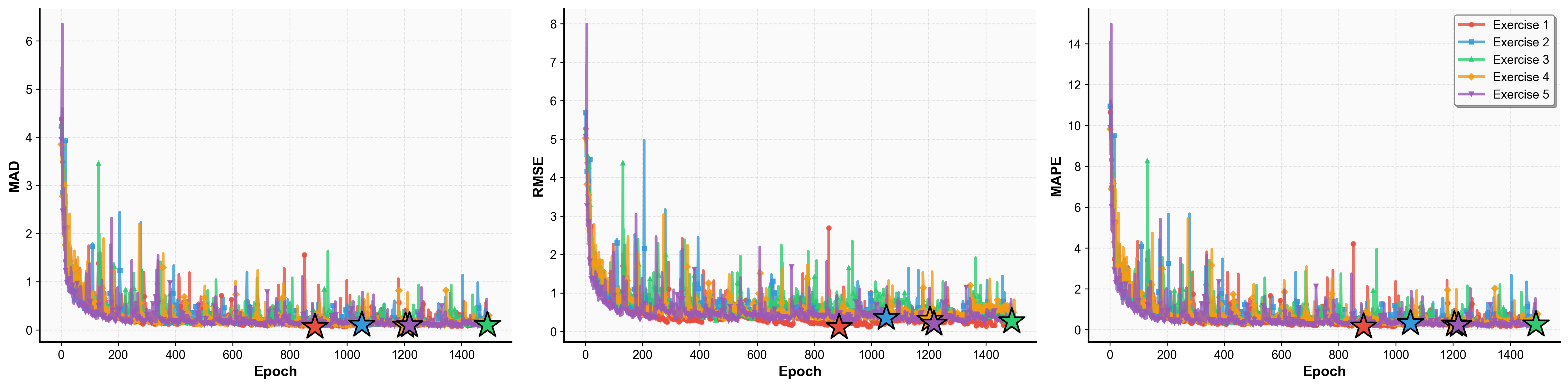}
    \caption{Training convergence curves for CC$^2$-STG on the KiMoRe dataset across Exercises 1-5 where left for MAD, enter for RMSE, and right for MAPE, and star markers indicating the epoch at which each exercise achieving optimal validation performance.}
    \label{fig:conv}
\end{figure*}

Figure~\ref{fig:conv} illustrates the training convergence of CC$^2$-STG across all five exercises in the KiMoRe dataset. All exercises exhibit rapid descent during the initial phase, followed by gradual refinement with diminishing oscillations, and Exercise 1 converges earliest at epoch 887 achieving the lowest MAD of 0.066 and RMSE of 0.118, while Exercise 3 requires the longest optimization, reaching optimal performance at epoch 1490. Exercises 2, 4, and 5 achieve convergence at epochs 1051, 1204, and 1218, respectively.

\subsection{Ablation Study}
We conduct ablation studies to validate both the contribution of each stream and the choice of distance-function pair in the geometric stream. We select Exercise~5 as the representative setting because it is the most challenging exercise in KiMoRe, exhibiting the largest variance among baseline methods and involving complex compensatory patterns, making it the most discriminative benchmark for component analysis. In the original KiMoRe study, both rule-based and template-based baselines exhibit their weakest correlation with clinical scores on this exercise~\citep{kimore}.

\begin{table}[!ht]
\centering
\caption{Ablation study of different stream configurations.}
\label{tab:ablation_streams}
\begin{tabular}{ccc|ccc}
\toprule
\textbf{S1} & \textbf{S2} & \textbf{S3}
& \textbf{MAD}~$\downarrow$ & \textbf{RMSE}~$\downarrow$ & \textbf{MAPE}~$\downarrow$ \\
\midrule
\cmark & \xmark & \xmark & 0.360 & 0.794 & 0.830 \\
\xmark & \cmark & \xmark & 2.977 & 3.334 & 7.368 \\
\xmark & \xmark & \cmark & 1.322 & 2.292 & 3.302 \\
\cmark & \cmark & \xmark & 3.642 & 4.654 & 8.947 \\
\cmark & \xmark & \cmark & 0.439 & 0.733 & 1.035 \\
\xmark & \cmark & \cmark & 3.889 & 5.105 & 9.892 \\
\cmark & \cmark & \cmark & \textbf{0.099} & \textbf{0.207} & \textbf{0.229} \\
\bottomrule
\end{tabular}
\end{table}

\noindent\textbf{Architecture streams.} The skeleton stream (S1) alone achieves MAD of 0.360, while CC stream (S2) and topological stream (S3) individually perform poorly with MAD of 2.977 and 1.322, respectively indicating that specialized features require complementary information. Two-stream combinations show mixed results, with S1+S3 achieving MAD of 0.439, while S1+S2 and S2+S3 perform worse than S1 alone. The full three-stream model yields MAD of 0.099, RMSE of 0.207, and MAPE of 0.229 representing 72.5\% improvement over the best single-stream configuration, confirming that synergistic integration of all three streams is essential for effective rehabilitation assessment.

\noindent\textbf{Distance-function pair.} To justify our choice of decomposition functions, we replace $(\phi_1, \phi_2) = (d^2, \log(d{+}1))$ with alternative transformations as outlined in Table~\ref{tab:ablation_function_pairs}. Single-function baselines perform worst, showing that one transformation alone cannot capture both fine-grained and global geometric variation, with the squared distance overemphasizing large deviations, while the log compresses the signal too aggressively. Paired transformations consistently outperform single ones, supporting the core hypothesis that complementary functions provide richer geometric information. Among pairs, the polynomial $d^2, d^3$ inherits the monotonic growth profile of $d^2$ alone, and the RBF pair $e^{-d}, e^{-d^2}$ saturates for large distances, attenuating the very deviations most diagnostic of pathological movements.

\begin{table}[!ht]
\centering
\caption{Ablation on the choice of distance-function pair.}
\label{tab:ablation_function_pairs}
\begin{tabular}{lccc}
\toprule
\textbf{Function pair} & \textbf{MAD}~$\downarrow$ & \textbf{RMSE}~$\downarrow$ & \textbf{MAPE}~$\downarrow$ \\
\midrule
  Linear only ($d$) & 0.182 & 0.323 & 0.399 \\
  Squared only ($d^2$) & 0.218 & 0.300 & 0.481 \\
  Log only ($\log(d+1)$) & 0.140 & 0.235 & 0.312 \\
  Polynomial pair ($d^2, d^3$) & 0.130 & 0.231 & 0.285 \\
  RBF pair ($e^{-d}, e^{-d^2}$) & 0.114 & 0.171 & 0.238 \\
  Sqrt + log ($\sqrt{d}, \log(d+1)$) & 0.155 & 0.276 & 0.335 \\
  \textbf{Ours} ($d^2, \log(d+1)$) & \textbf{0.099} & \textbf{0.207} & \textbf{0.229} \\
\bottomrule
\end{tabular}
\end{table}

\section{Conclusion}

In this paper, we present CC$^2$-STG, a three-stream framework for skeleton-based clinical assessment that combines a convex-concave decomposition of pairwise joint distances, sparse proximal feature selection, and adaptive multi-scale $k$-NN topological modeling. Our method achieves state-of-the-art performance on KiMoRe across all exercises and metrics, competitive results on UIPRMD validating cross-dataset generalization, and interpretable joint-pair selection through soft-thresholded sparse weights. Our framework can also be extended to other quality-assessment tasks beyond rehabilitation.

\noindent\textbf{Limitations and Future Work.} Our reliance on per-frame $k$-NN topology inherits the sensitivity of nearest-neighbor graphs to noisy joint coordinates. We plan to address these by extending CC$^2$-STG with multi-modal depth and RGB fusion, developing interpretability tools that visualize the learned sparse weights for therapist use, and validating on broader patient cohorts spanning additional pathologies.

\section*{Author Contributions}
Q.-A.N.D. and M.H.H. conceived the study and developed the core methodology. Q.-A.N.D. and D.P.M. implemented the software framework and conducted the experimental evaluations. M.-A.N. and H.Q.K. performed data curation and formal analysis. M.L. and O.T.-C.C. provided technical supervision and validation. Q.-A.N.D., D.P.M., and M.-A.N. wrote the main manuscript text and prepared the figures. M.H.H. and O.T.-C.C. supervised the project and managed administration. All authors reviewed and approved the final manuscript.

\section*{Funding Declaration}
This work have no funding.

\bibliography{sn-bibliography}

\newpage
\section*{Appendix A: Evaluation Metrics}

We evaluated the CC$^2$-STG model using three standard metrics of RMSE, MAD, and MAPE metrics as expressed in Eq~\eqref{m3} - Eq~\eqref{m1}.

\begin{equation}
\text{RMSE} = \sqrt{\frac{1}{n} \sum_{i=1}^{n} (y - \hat{y})^2}
\label{m3}
\end{equation}
\begin{equation}
\text{MAD} = \frac{1}{n} \sum_{i=1}^{n} |y - \hat{y}|
\label{m1}
\end{equation}
\begin{equation}
\text{MAPE} = \frac{1}{n} \sum_{i=1}^{n} \frac{|y - \hat{y}|}{y} \times 100
\label{m2}
\end{equation}
where $y$ represents an actual value, $\hat{y}$ denotes a predictions, and $n$ is the sample size. RMSE emphasizes larger errors through squaring, making it sensitive to significant deviations in prediction precision. MAD treats all errors equally by measuring average absolute differences without amplification. MAPE expresses errors as percentages of actual values, enabling comparison across exercises with different motion scales.

\section*{Appendix B: Proof of Lemma 1}
\begin{proof}
Let $x$ denote the pairwise distance between joints, i.e., $x = d_{ij}^t = \|\mathbf{p}_{t,i} - \mathbf{p}_{t,j}\|_2$ for joints $i$ and $j$ at time $t$.

For $\phi_1(x) = x^2$, the first derivative is $\phi_1'(x) = 2x$ and the second derivative is $\phi_1''(x) = 2 > 0$, establishing strict convexity. The superlinear growth follows from:
\begin{equation}
\lim_{x \to \infty} \frac{\phi_1(x)}{x} = \lim_{x \to \infty} \frac{x^2}{x} = \lim_{x \to \infty} x = \infty.
\end{equation}

For $\phi_2(x) = \log(x+1)$, the first derivative is $\phi_2'(x) = \frac{1}{x+1}$ and the second derivative is $\phi_2''(x) = -\frac{1}{(x+1)^2} < 0$, establishing strict concavity. The sublinear growth follows from L'H\^{o}pital's rule:
\begin{equation}
\lim_{x \to \infty} \frac{\phi_2(x)}{x} = \lim_{x \to \infty} \frac{\log(x+1)}{x} = \lim_{x \to \infty} \frac{1}{x+1} = 0.
\end{equation}

The complementarity of these functions-quadratic amplification versus logarithmic compression-ensures that $\phi_1$ emphasizes large-scale structural deviations while $\phi_2$ captures fine-grained local relationships. 
\end{proof}

\section*{Appendix C: Proof of Proposition 1}
\begin{proof}
For any $d_a \neq d_b > 0$, the ratio $\phi_1(d_a)/\phi_1(d_b) = (d_a/d_b)^2$ is governed by a quadratic scaling, whereas 

\begin{equation}
    \frac{\phi_2(d_a)}{\phi_2(d_b)} = \frac{\log(d_a+1)}{\log(d_b+1)}
\end{equation}

follows a logarithmic scaling. Since $(d_a/d_b)^2 \neq \log(d_a+1)/\log(d_b+1)$ in general, the two functions impose distinct orderings on pairwise distances, confirming that their concatenation captures geometric information at complementary scales.
\end{proof}

\section*{Appendix D: Proof of Theorem 1}
\begin{proof}
The following restates the standard proximal gradient convergence result~\citep{beck2009fast} applied to the elastic-net objective in Eq.~\eqref{eq:tls}, included here for completeness and to verify the step-size condition relevant to our setting. The proximal gradient method for the composite objective $\mathcal{L}(\mathbf{w}) = f(\mathbf{w}) + g(\mathbf{w})$ with $f = \mathcal{L}_{\text{task}} + \lambda\|\mathbf{w}\|_2^2$ (smooth part) and $g = \mu\|\mathbf{w}\|_1$ (non-smooth part) takes the form~\citep{parikh2014proximal}:
\begin{equation}
\mathbf{w}^{(k+1)} = \text{prox}_{\eta g}\left(\mathbf{w}^{(k)} - \eta \nabla f(\mathbf{w}^{(k)})\right),
\end{equation}
where the proximal operator of $g$ is defined as:
\begin{equation}
\text{prox}_{\eta g}(\mathbf{z}) = \arg\min_{\mathbf{w}} \left\{\frac{1}{2\eta}\|\mathbf{w} - \mathbf{z}\|_2^2 + \mu\|\mathbf{w}\|_1\right\}.
\end{equation}
This proximal operator has a closed-form solution given by the soft thresholding operator~\citep{donoho1995noising} with $\text{prox}_{\eta g}(\mathbf{z}) = \mathcal{S}_{\eta\mu}(\mathbf{z})$. For the gradient step, we have
\begin{equation}
\nabla f(\mathbf{w}) = \nabla \mathcal{L}_{\text{task}}(\mathbf{w}) + 2\lambda\mathbf{w}
\end{equation}
substituting into the proximal update as follow:
\begin{align}
\mathbf{w}^{(k+1)} &= \mathcal{S}_{\eta\mu}\left(\mathbf{w}^{(k)} - \eta(\nabla \mathcal{L}_{\text{task}}(\mathbf{w}^{(k)}) + 2\lambda\mathbf{w}^{(k)})\right) \\
&= \mathcal{S}_{\eta\mu}\left((1-2\lambda\eta)\mathbf{w}^{(k)} - \eta\nabla \mathcal{L}_{\text{task}}(\mathbf{w}^{(k)})\right).
\end{align}
substituting into the proximal update gives:
\begin{align}
\mathbf{w}^{(k+1)} 
&= \mathcal{S}_{\eta\mu}\!\left(\mathbf{w}^{(k)} - \eta\bigl(\nabla 
\mathcal{L}_{\text{task}}(\mathbf{w}^{(k)}) + 2\lambda\mathbf{w}^{(k)}\bigr)\right) \\
&= \mathcal{S}_{\mu\eta}\!\left((1-2\lambda\eta)\mathbf{w}^{(k)} 
- \eta\nabla \mathcal{L}_{\text{task}}(\mathbf{w}^{(k)})\right),
\end{align}
which is exactly the update in Theorem~1 with threshold $\mu\eta$.

Convergence follows from standard proximal gradient convergence theory of~\citep{beck2009fast}, since $f$ is $(L+2\lambda)$-smooth and $g$ is convex (though non-smooth). The sequence $\{\mathbf{w}^{(k)}\}$ converges to a stationary point when $\eta < \frac{1}{L + 2\lambda}$, ensuring that the descent property $\mathcal{L}(\mathbf{w}^{(k+1)}) \leq \mathcal{L}(\mathbf{w}^{(k)})$ holds.
\end{proof}

\section*{Appendix E: Proof of Lemma 2}
\begin{proof}
Based on the triangle inequality, for any two joints $i$ and $j$:
\begin{equation}
\begin{aligned}
|d_{ij}^t - d_{ij}'^t| &= \left|\|\mathbf{p}_{t,i} - \mathbf{p}_{t,j}\|_2 - \|\mathbf{p}_{t,i}' - \mathbf{p}_{t,j}'\|_2\right| \\
&\leq \|(\mathbf{p}_{t,i} - \mathbf{p}_{t,j}) - (\mathbf{p}_{t,i}' - \mathbf{p}_{t,j}')\|_2 \\
&\leq \|\mathbf{p}_{t,i} - \mathbf{p}_{t,i}'\|_2 + \|\mathbf{p}_{t,j} - \mathbf{p}_{t,j}'\|_2 < 2\delta.
\end{aligned}    
\end{equation}

The mean feature is
\begin{equation}
\begin{aligned}
|\mu_k^{(i)}(\mathbf{P}_t) - \mu_k^{(i)}(\mathbf{P}_t')|
&= \left|\frac{1}{k}\sum_{j \in \mathcal{N}_k(i)} (d_{ij}^t - d_{ij}'^t)\right| \\
&\le \frac{1}{k}\sum_{j \in \mathcal{N}_k(i)} |d_{ij}^t - d_{ij}'^t|
< \frac{1}{k} \cdot k \cdot 2\delta \\
&= 2\delta.
\end{aligned}
\end{equation}

For the minimum distance, since $|\min_j a_j - \min_j b_j| \leq \max_j |a_j - b_j|$ for any sequences $\{a_j\}$ and $\{b_j\}$, we have:
\begin{equation}
|\delta_k^{(i)}(\mathbf{P}_t) - \delta_k^{(i)}(\mathbf{P}'_t)| \leq \max_{j \in \mathcal{N}_k(i)} |d_{ij}^t - d_{ij}'^t| < 2\delta.
\end{equation}
This stability property ensures that small perturbations in joint positions do not drastically change the topological features, providing robustness for clinical assessment.
\end{proof}

\section*{Appendix F: Proof of Theorem 2}
\begin{proof}
The equality of k-NN features across all scales implies:
\begin{enumerate}
    \item Identical mean distances: $\sum_{j \in \mathcal{N}_k(i)} d_{ij}^t = \sum_{j \in \mathcal{N}_k(i)} d_{ij}'^t$ for all $i$ and $k$
    \item Identical standard deviations: $\sigma_k^{(i)}(\mathbf{P}_t) = \sigma_k^{(i)}(\mathbf{P}'_t)$ for all $i$ and $k$
    \item Identical minimum distances: $\min_{j \in \mathcal{N}_k(i)} d_{ij}^t = \min_{j \in \mathcal{N}_k(i)} d_{ij}'^t$
\end{enumerate}

Together, these constraints ensure that the local distance structure within $k$-neighborhoods is identical between the two configurations. While this does not guarantee global isometry (which would require all $\binom{N}{2}$ pairwise distances to be matched), it makes that the local coordination patterns between nearby joints are preserved.

For clinical movement quality assessment, local joint coordination is more diagnostically relevant than global body configuration~\citep{bernstein1967coordination}, making this local preservation property sufficient and appropriate for the rehabilitation exercise evaluation task.
\end{proof}

\begin{figure}[!b]
    \centering
    \includegraphics[width=0.6\linewidth]{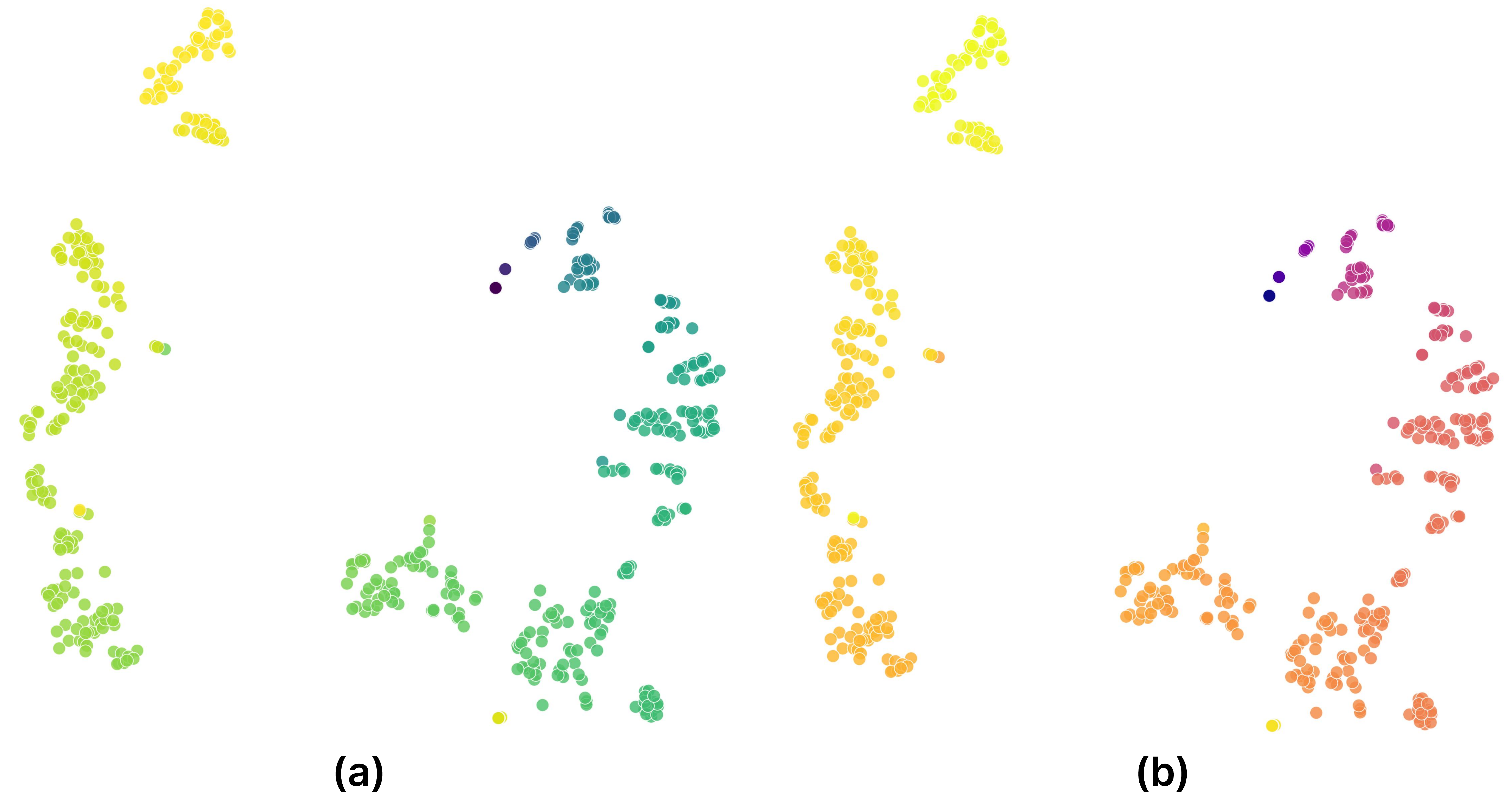}
    \caption{t-SNE visualization of feature representations from CC$^2$-STG model. (a) Ground truth clinical score distribution. (b) Learned feature space from CC$^2$-STG.}
    \label{fig:te}
\end{figure}

\section*{Appendix G: Tsne Visualizations}
Figure~\ref{fig:te} presents t-SNE visualizations comparing the ground truth score distributions with the learned feature representations from CC$^2$-STG. The ground truth embedding in Figure \ref{fig:te} (a) shows the natural clustering structure of clinical assessment scores, while the proposed model's feature space in Figure \ref{fig:te} (b) demonstrates that our learned representations preserve the inherent structure. The similar clustering patterns between (a) and (b) indicate that CC$^2$-STG effectively captures the underlying clinical score distribution, with distinct clusters corresponding to different quality levels of rehabilitation movements. The proposed model preserves the inherent clustering structure of clinical assessments, demonstrating effective representation learning for movement quality discrimination.

\section*{Appendix H: Joint Important Visualizations} \label{iii}

\begin{figure}[!ht]
    \centering
    \includegraphics[width=\columnwidth]{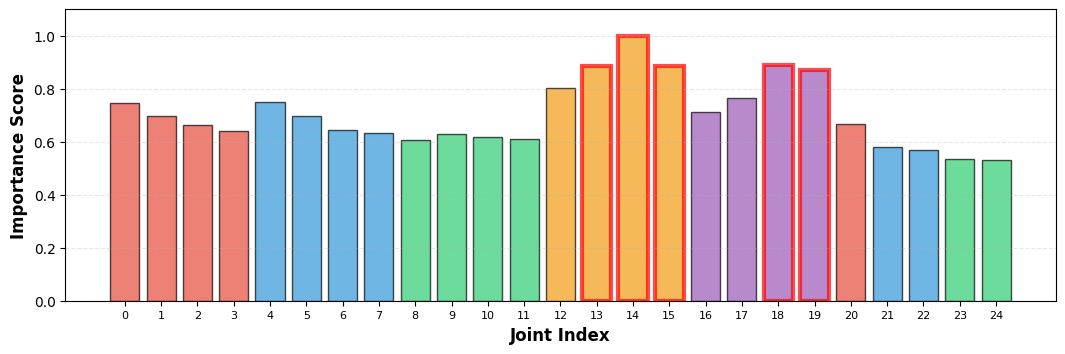}
    \includegraphics[width=\columnwidth]{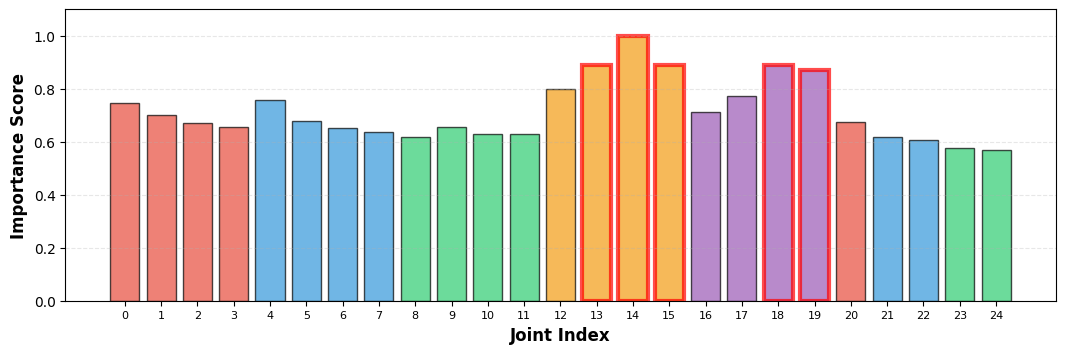}
    \includegraphics[width=\columnwidth]{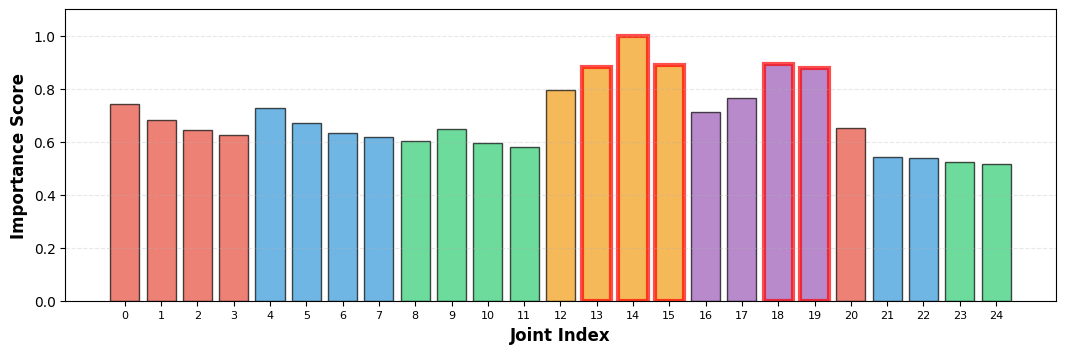}
    \includegraphics[width=\columnwidth]{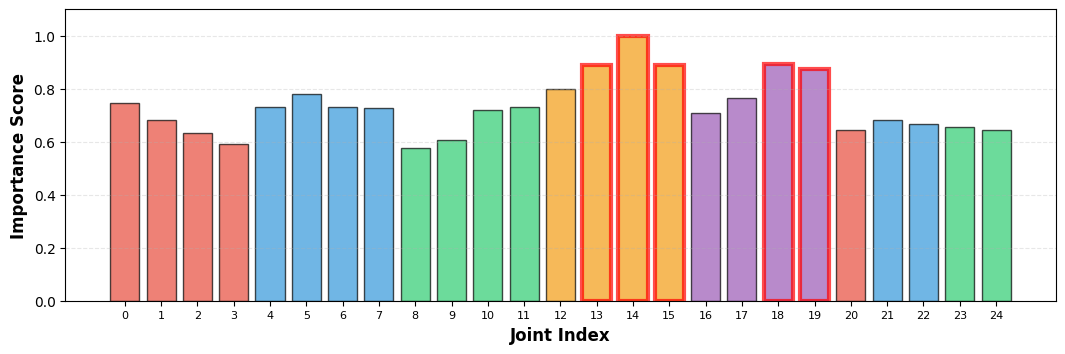}
    \includegraphics[width=\columnwidth]{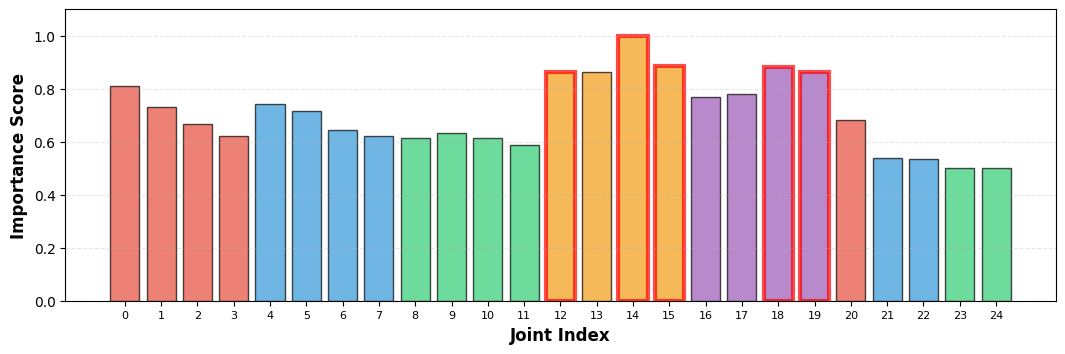}
    \caption{Learned joint importance scores from CC$^2$-STG across KiMoRe Exercises 1-5 (top to bottom), where joints with highlighted borders indicating the highest importance weights.}
    \label{fig:convergence}
\end{figure}

Figure~\ref{fig:convergence} visualizes the learned joint importance scores from the sparse feature weighting mechanism in the Convex-Concave geometric stream across all five exercises. Across all exercises, joint 14 consistently receives the highest importance weight, followed by joints 12, 13, 15, 18, and 19 (highlighted with red borders). These joints correspond to the hip and lower limb regions, which are clinically significant for assessing balance, coordination, and compensatory movements in rehabilitation exercises. Notably, peripheral joints (21-24) consistently receive lower importance scores, indicating that the model learning approaches to focus on core body structures most relevant to movement quality assessment.

\end{document}